\def\year{2017}\relax
\documentclass[letterpaper]{article}
\usepackage{aaai17}
\usepackage{times}
\usepackage{helvet}
\usepackage{courier}
\usepackage{graphicx}
\usepackage{caption}
\usepackage{subcaption}
\usepackage{amsmath}
\usepackage{algorithm}
\usepackage{algorithmic}

\renewcommand{\vec}[1]{\boldsymbol#1}
\newcommand{\word}[1]{\texttt{#1}}
\newcommand{\sysname}[1]{\textsc{#1}}

\newcommand{\citet}[1]{\citeauthor{#1} \shortcite{#1}}

\begin{document}
%
\title{Topic Browsing for Research Papers with Hierarchical Latent Tree Analysis}
\author{Leonard K.M. Poon\\
The Education University of Hong Kong\\
10 Lo Ping Road, Hong Kong, China\\
kmpoon@eduhk.hk
\And
Nevin L. Zhang \\
The Hong Kong University of Science and Technology \\
Clear Water Bay Road, Hong Kong, China \\
lzhang@cse.ust.hk
}
\maketitle
\begin{abstract}
Academic researchers often need to face with a large collection of research papers in the literature.  This problem may be even worse for postgraduate students who are new to a field and may not know where to start.  To address this problem, we have developed an online catalog of research papers where the papers have been automatically categorized by a topic model.  The catalog contains 7719 papers from the proceedings of two artificial intelligence conferences from 2000 to 2015. Rather than the commonly used Latent Dirichlet Allocation, we use a recently proposed method called hierarchical latent tree analysis for topic modeling.  The resulting topic model contains a hierarchy of topics so that users can browse the topics from the top level to the bottom level.  The topic model contains a manageable number of general topics at the top level and allows thousands of fine-grained topics at the bottom level.  It also can detect topics that have emerged recently.
\end{abstract}

\section{Introduction}

Academic researchers often need to face with a large collection of research papers in the literature.  This problem may be even worse for postgraduate students who are new to a field and may not know where to start.  Researchers usually use keywords to search for related papers using a search engine.  After reading some papers, they then try to group related papers together to discover the main topics in the field.  This process can be time-consuming.

The approach above can be regarded as bottom-up approach.  A top-down approach would be to start with topic hierarchy.  Researchers can then pick  a general topic and drill down to more specific topics.  Papers related to any of the topics can be presented to the researchers when requested.

To allow the top-down approach, traditionally a taxonomy has to be defined manually.  Papers can then be manually categorized according to the taxonomy.  One problem with the traditional method is that it requires much effort.  Besides, the topics in the taxonomy may not be able to keep up with recent development.

Topic models can be used to automate this process.  They can be used to detect topics from a collection of documents and categorize documents according to the detected topics.  We use a recently proposed method called \emph{hierarchical latent tree analysis} (HLTA)~\cite{liu14hierarchical,chen16progressive} for topic modeling.  Unlike Latent Dirichlet Allocation~\cite{blei03latent}, HLTA yields a hierarchy of topics.  hLDA~\cite{blei10nested} and nHDP~\cite{paisley15nested} are two extensions of LDA for producing topic hierarchy.  However, HLTA has been recently shown to produce better quality of topics and topic hierarchy than the two LDA extensions~\cite{chen16progressive}. 

To facilitate the top-down approach, we have developed an online catalog of research papers where the papers have been automatically categorized by a topic model built with HLTA.  The catalog contains 7719 papers from the proceedings of two artificial intelligence conferences from 2000 to 2015.  The resulting topic model contains a hierarchy of topics so that users can browse the topics from the top level to the bottom level.  The topic model contains a manageable number of general topics at the top level and allows thousands of fine-grained topics at the bottom level.  It also can detect topics that have emerged recently.

This paper is organized as follows.  In the next section we review background of our work.  We then explain HLTA using an example.  Next, we describe the procedure for building the online catalog.  In the following section, we show the results and observations obtained from the online catalog. After that, we conclude the paper.

%
%

\section{Background}


In topic modeling, documents are usually represented as \emph{bags of words}.  Consider a collection $\mathcal{D}=\{d_1, \ldots, d_N\}$ of $N$ documents .  Suppose $M$ words are included in the vocabulary $\mathcal{V}=\{w_1, \ldots, w_M\}$.  Each document $d$ can be represented as a vector $d=(c_1, \ldots, c_M)$, where $c_i$ represents the count of word $w_i$ occurring in the document $d$.  The aim of topic modeling is to detect a number $K$ of topics $z_1, \ldots, z_K$ among the documents $\mathcal{D}$.  The number $K$ can be given or learned.  The topic model defines a distribution over words for each topic.  A topic is often characterized by representative words based on the distribution.

\emph{Latent Dirichlet Allocation} (LDA) is a popular method for topic modeling~\cite{blei03latent,blei12probabilistic}.  LDA assumes each document $d$ to belong to the $K$ topics according to a distribution $P(Z=z_k|d)$ over the topics.  In other words, $\sum_1^K P(z_k|d) = 1$.  This kind of model is known as \emph{mixed membership model}.  For each topic $z_k$, LDA defines a conditional distribution $P(w_i|z_k)$ over words $w_i$.  A topic $z_k$ can then be characterized by the most probable words according to $P(w_i|z_k)$.

\subsection{Latent Tree Models}

A \emph{latent tree model} (LTM) is a tree-structured probabilistic graphical model~\cite{zhang04hierarchical,chen12model}.  Figure~\ref{fig:z344-model} shows an example of LTM.  When an LTM is used for topic modeling, the leaf nodes represent the observed word variables $\vec{W}$, whereas the internal nodes represent the unobserved topic variables $\vec{Z}$.   All variables are binary.  Each word variable $W_i \in \vec{W}$ indicates the presence or absence of the word $w_i \in \mathcal{V}$ in a document.  Each topic variable $Z_i \in \vec{Z}$ indicates whether a document belongs to the i-th topic.

For technical convenience, we often root an LTM at one of its latent nodes and regard it as a Bayesian network~\cite{pearl88probabilistic}. Then all the edges are directed away from the root. The numerical information of the model includes a marginal distribution for the root and one conditional distribution for each edge. For example, edge $Z1314$ $\rightarrow$ \word{dean} is associated with probability $P(\word{dean} | Z1314)$. The conditional distribution associated with each edge characterizes the probabilistic dependence between the two nodes that the edge connects. The product of all those distributions defines a joint distribution over all the latent variables $\vec{Z}$ and observed variables $\vec{W}$.  Denote the parent of a variable $X$ as $pa(X)$ and let $pa(X)$ be an empty set when $X$ is the root. Then the LTM defines a joint distribution over all observed and latent variables as follows:
$$P(\vec{W},\vec{Z}) = \prod_{X\in \vec{W}\cup\vec{Z}} P(X|pa(X))$$.

Given a document $d$, the values of word variables $\vec{W}$ are observed.  Use $d = (w_1, \ldots,w_M)$ to denote also those observed values.  Whether a document $d$ belongs to a topic $Z \in \vec{Z}$ can be determined by the probability $P(Z|d)$.  The LTM gives a \emph{multi-membership model} since a document can belong to multiple topics.  Unlike in LDA, the topic probabilities $P({Z}|d)$ in LTM do not necessarily sum to one.

\begin{figure}
\centering
\begin{subfigure}[b]{0.49\textwidth}
\includegraphics[width=\textwidth]{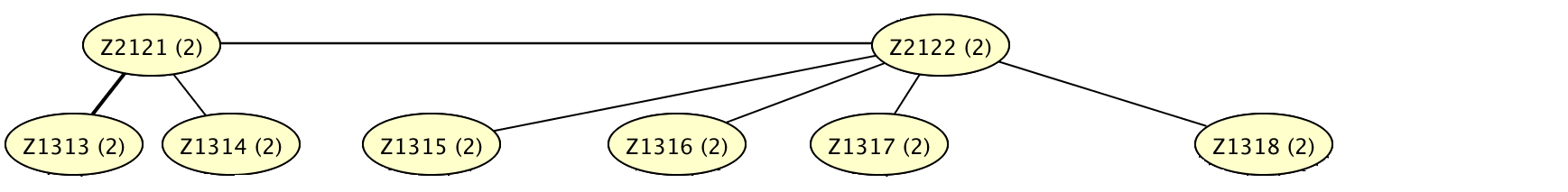}
\caption{Finding level-2 latent variables}
\label{fig:hlta-find-2}
\end{subfigure}
\begin{subfigure}[b]{0.49\textwidth}
\vspace{0.5em}
\includegraphics[width=\columnwidth]{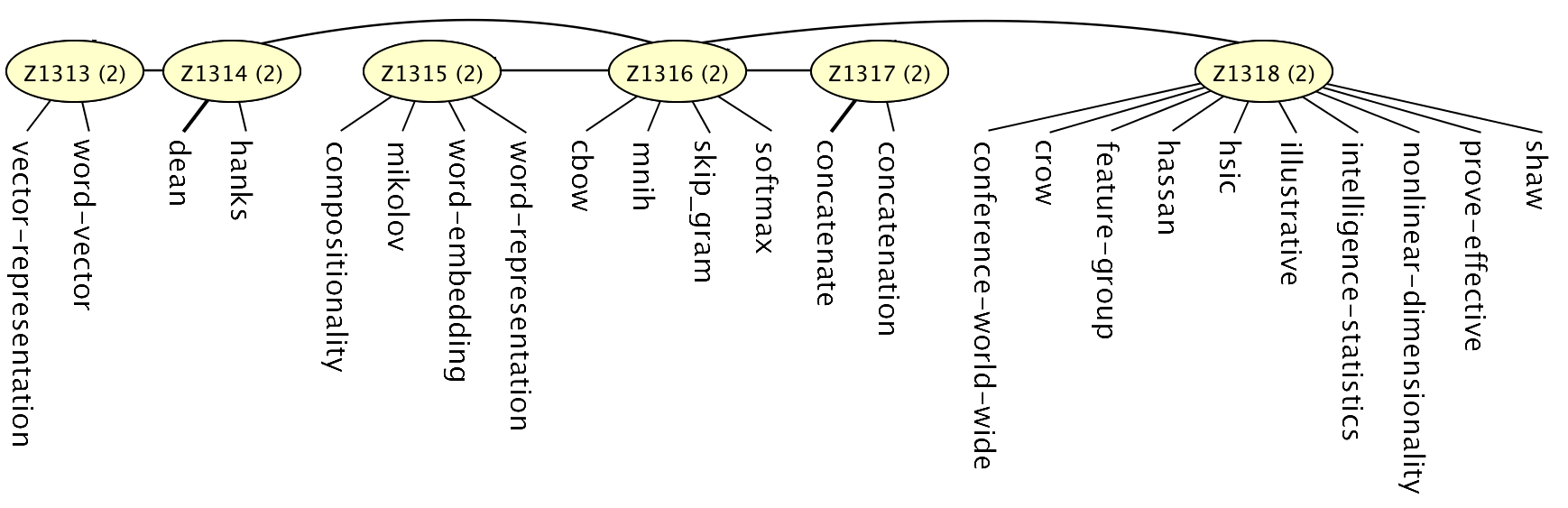}
\caption{Linking level-1 latent variables}
\label{fig:hlta-link-1}
\end{subfigure}
\begin{subfigure}[b]{0.49\textwidth}
\vspace{0.5em}
\includegraphics[width=\columnwidth]{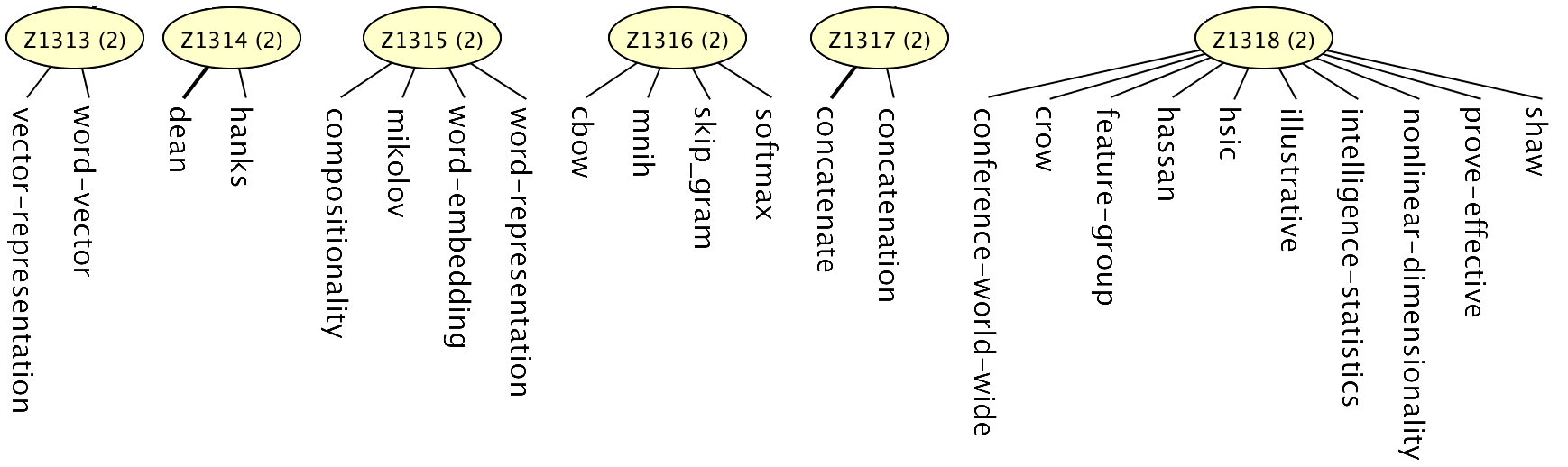}
\caption{Finding level-1 latent variables}
\label{fig:hlta-find-1}
\end{subfigure}
\caption{Illustrative example of \sysname{PEM-HLTA}.}
\end{figure}

\section{Hierarchical Latent Tree Analysis}

For topic modeling, an LTM has to be learned from the document data $\mathcal{D}$.  This requires learning the number of topic variables, the connection between the variables, and the probabilities in the model.

We use the method \sysname{PEM-HLTA} proposed by~\citet{chen16progressive} to build LTMs for topic modeling.  The method builds LTMs level by level and is thus also known as \emph{hierarchical latent tree analysis} (HLTA).  In this section, we use an example to illustrate the main ideas of \sysname{PEM-HLTA}.  Readers are referred to the original paper for details.

As an example, consider a data set that contains the 24 word variables in Figure~\ref{fig:z344-model}.  \sysname{PEM-HLTA} is an iterative procedure that builds one level of model in each iteration.  In the first iteration, it partitions the 24 word variables in 6 clusters (Figure~\ref{fig:hlta-find-1}).  The clusters are \emph{unidimensional} in the sense that the co-occurrences of words in each cluster can be properly modeled using a single latent variable.  A latent variable is introduced for each cluster to form model in which all variables in a cluster are connected to the latent variable.  We metaphorically refer to those models corresponding to the clusters as islands and the latent variables in them as level-1 latent variables.

The next step is to link up the 6 islands. This is done by estimating the mutual information (MI)~\cite{cover06elements} between every pair of latent variables and building a Chow-Liu tree~\cite{chow68approximating} over them, so as to form an overall model~\cite{liu13greedy}. The result is the model in the Figure~\ref{fig:hlta-link-1}.

To build the next level of model, inference is carried out to compute the posterior distribution of each level-1 latent variable for each document.  The document is assigned to the state with the maximum posterior probability, resulting in a data set over the level-1 latent variables.  The data set is used as the input for the next iteration.  In the second iteration, the level-1 latent variables are partitioned into 2 groups. The 2 islands are linked up to form the model shown in Figure~\ref{fig:hlta-find-2}.  The model in Figure~\ref{fig:hlta-find-2} is then stacked on the model in Figure~\ref{fig:hlta-link-1} and a new data set over level-2 latent variables are computed.  The iteration continues until the model in Figure~\ref{fig:z344-model} is obtained.

Intuitively, the co-occurrence of words are captured by the level-1 latent variables, whose co-occurring patterns are captured by higher level latent variables. Then a topic hierarchy can be extracted, with topics on top more general and topics at the bottom more specific.

\section{Building Online Catalog with HLTA}

In this section, we describe the procedure for building an online catalog of documents with HLTA.  The procedure starts with each document contained in a PDF file.

\subsection{Extract Text}

Given a PDF file, we extract the text content using Apache PDFBox.\footnote{http://pdfbox.apache.org/}  We remove hyphenation from the extracted text.  After that we use Stanford Core NLP~\cite{manning14stanford} for sentence splitting and lemmatization.\footnote{http://stanfordnlp.github.io/CoreNLP/}

The normalize the words, we convert all letters to lowercase.  We also remove  accents and ligatures using the \texttt{java.text.Normalizer} class in the Java library.  We use underscore to replace all non-alphanumeric characters and starting digits in a word.  We remove stop words\footnote{http://jmlr.csail.mit.edu/papers/volume5/lewis04a/a11-smart-stop-list/english.stop} and words with fewer than 4 characters.

\begin{algorithm}[t]
\caption{Find-NGrams$(\mathcal{D}, m, M)$}
\label{algo:find-ngrams}

\textbf {Input}:  $\mathcal{D}$ -- Document collection, $m$ -- maximum value of $n$, $M$ -- number of tokens to be selected.\\
\textbf {Output}: A document collection with individual words replaced by selected n-grams. \\
\vspace{-0.5em}

\begin{algorithmic}[1]
\STATE Find individual words for each $d\in\mathcal{D}$.
\STATE Set $\mathcal{V}$ to be the set of the $M$ selected words.
\FOR{$n = 2$ to $m$}
\STATE For each $d\in\mathcal{D}$, form an $n$-gram for each pair of consecutive tokens $t_1$, $t_2$ in $d$ if $t_1,t_2 \in \mathcal{V}$ and if the resulting $n$-gram has a length of $n$.
\STATE Denote the set of newly formed $n$-grams by $\mathcal{U}$.
\STATE Set $\mathcal{V}$ to be the set of $M$ tokens $t \in \mathcal{V} \cup \mathcal{U}$ that have the highest $\textrm{tf-idf}(t)$.
\STATE For each $d\in\mathcal{D}$, replace all pairs of consecutive tokens that can be used to form an $n$-gram in $\mathcal{V}$.  Let $\mathcal{D}$ be the collection of documents after replacement.
\ENDFOR
\STATE {\bf return} $\mathcal{D}$.

\end{algorithmic}
\end{algorithm}

\subsection{Convert Data}

After text is extracted, we compute the term frequency and document frequency.  The term frequency $\textrm{tf}(w,d)$ is defined as the number of occurrences of a word $w$ in document $d$.  The document frequency $\textrm{df}(w)$ is defined as the number of documents that contain the word $w$.  

We remove words that occur in more than 25\% of documents.  In other words, a word $w$ is removed if $\textrm{df}(w) \ge 0.25N$, where $N$ is the number of documents.  Given a number $M$, we select the $M$ words with highest TF-IDF, which is given by:
$$\textrm{tf-idf}(w) = \frac{1}{\ln \textrm{df}(w)} \sum_{d\in\mathcal{D}}\textrm{tf}(w,d).$$
After selecting the words, we can represent each document $d$ as a vector $d=(w_1, \ldots, w_M)$, where $w_i$ corresponds to one of the selected words and its value indicates the presence ($w_i=1$) or absence ($w_i=0$) of the word in $d$.

\begin{figure}
    \centering
    \begin{subfigure}[b]{0.49\textwidth}
        \includegraphics[width=\textwidth]{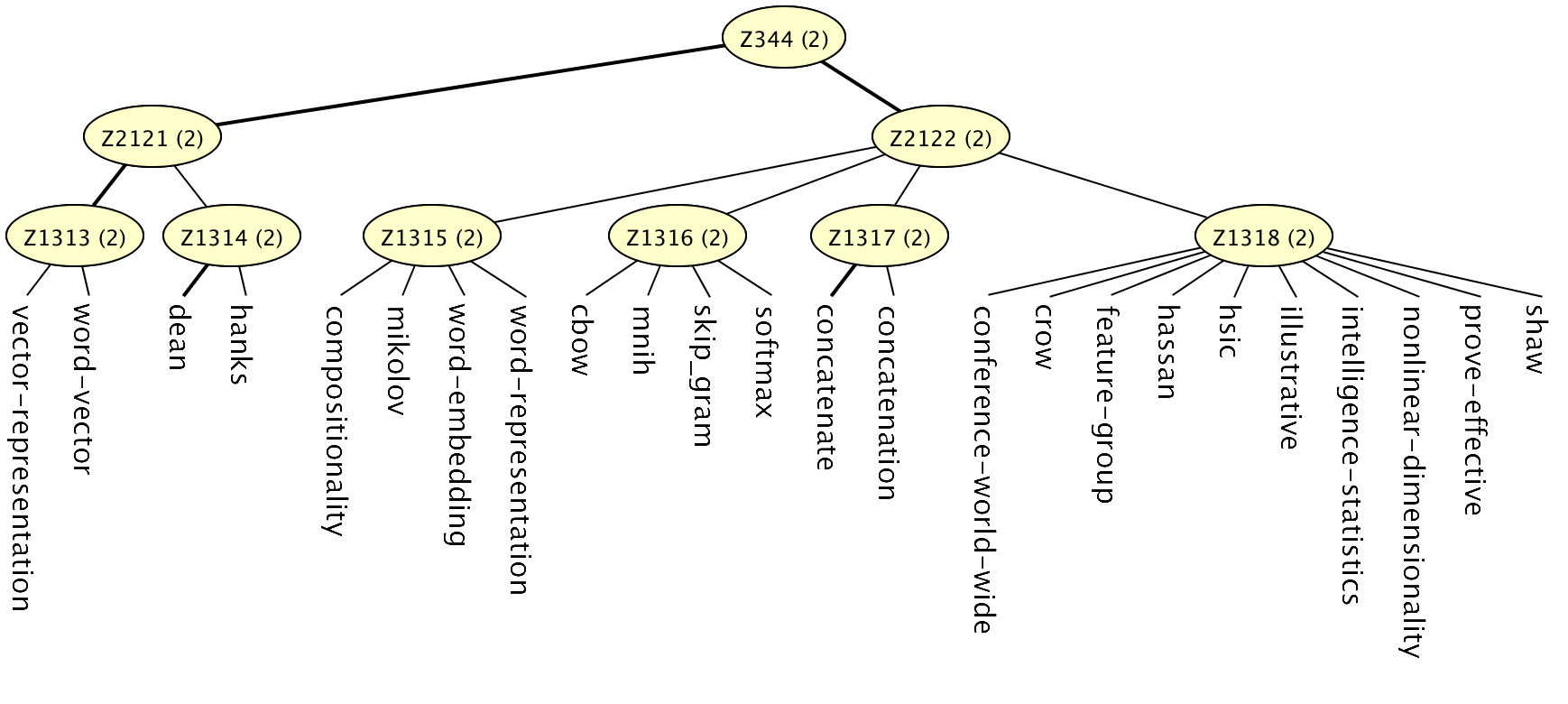}
        \caption{Latent tree model}
        \label{fig:z344-model}
    \end{subfigure}
    \\\vspace{0.5em}
    \begin{subfigure}[b]{0.49\textwidth}
        \includegraphics[width=\textwidth]{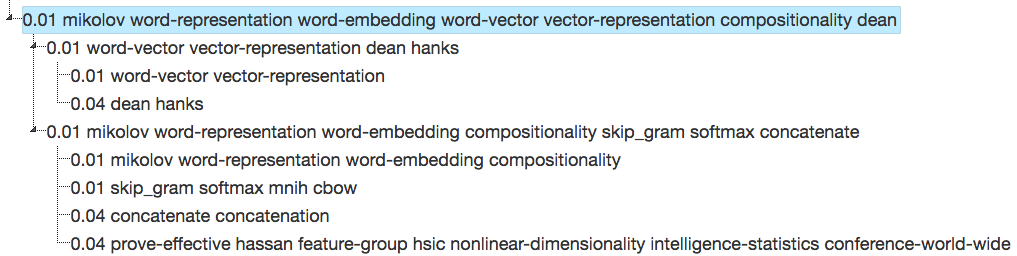}
        \caption{Topic hierarchy}
        \label{fig:z344-tree}
    \end{subfigure}
    \caption{Illustration of topic extraction.  Each shaded node in (a) corresponds to a row in (b).  Both of them represent a topic.  The topic hierarchy (b) is extracted from the model (a).}\label{fig:extraction}
\end{figure}

Inspired by~\citet{deng2016unsupervised}, we consider also n-grams in addition to individual words.  For a given value of $m$, we consider 1-grams, 2-grams, and so on up to $m$-grams.  We also use the term \emph{tokens} to refer to the n-grams.  The leaf nodes of the model in Figure~\ref{fig:z344-model} show some examples of n-grams.  The examples of 2-grams include \word{vector-representation} and \word{word-vector}, and an example of 3-gram is \word{conference-world-wide}.

The n-grams in a collection of documents $\mathcal{D}$ can be found by Algorithm~\ref{algo:find-ngrams}.  After running the algorithm, each document can be converted to a binary vector similarly as above.

The tree structure of LTM limits a word variable to be connected to exactly one parent.  The inclusion of n-grams mitigates this limitation.  For example, the word \word{bayesian} may be related to statistics and the word \word{network} to social networks or communities.  The 2-gram \word{bayesian-network} means a class of graphical models and has a different meaning from the individual words.  By including 2-grams, the three tokens are allowed to three different parent nodes that are more closely to their respective meanings.

\subsection{Building Model}

After converting the documents to binary vectors, we run \sysname{PEM-HLTA} using those vectors as input data.  Note that \sysname{PEM-HLTA} automatically determines the number of levels of latent variables and the number of latent variables in each level.  The result of \sysname{PEM-HLTA} is an LTM.

\subsection{Extract Topic Hierarchy}

An LTM defines a tree structure of nodes.  Each internal node represents a topic.  Since the model defines a joint distribution over the variables, we can compute the MI between every pair of variables.  For each topic, we compute the MI between each descendent word variable and the topic variable.  We then pick the descendent words with highest MI to characterize the topic.

As an example, consider extracting topics from the model in Figure~\ref{fig:z344-model}.  Each shaded node represents a topic.  Suppose we want to characterize $Z344$.  The descendent word variables are \word{vector-representation}, \word{word-vector}, $\ldots$, \word{shaw}.  We compute the MI between $Z344$ and each of those word variables.  The 7 words with highest MI are shown in the shaded row in Figure~\ref{fig:z344-tree}.  The row corresponds to the topic represented by $Z344$.  The second and third rows correspond to the topics represented by $Z2121$ and $Z1313$ respectively.  The topic extraction from the model in Figure~\ref{fig:z344-model} results in the topic hierarchy shown in Figure~\ref{fig:z344-tree}.

In addition to finding the characterizing words, we also estimate the size of each topic $Z$, which is given by the marginal probability $P(Z)$.  It shows how often a topic is estimated to occur in a document collection.  In Figure~\ref{fig:z344-tree}, the number on each row indicates the topic size.

\subsection{Build Online Catalog}

The LTM can be used to classify the documents according to the topics detected.  A document $d$ is assigned to topic $Z$ if $P(Z=1|d) > 0.5$.  We use a webpage to display this information.  On the webpage, a topic hierarchy similar to Figure~\ref{fig:z344-tree} is shown.  The hierarchy is built with a jquery plugin called jsTree.\footnote{https://www.jstree.com}  When a topic is clicked, a list of documents belonging to that topic will be shown.

\section{Results and Observations}

We have built an online catalog of research papers using the method discussed above.  The papers were obtained from the proceedings of two AI conferences, namely AAAI Conference on Artificial Intelligence and International Joint Conference on Artificial Intelligence.  Proceedings between year 2000 and 2015 were used.  The resulting collection contains 7719 papers.  We considered n-grams for $n=1,2,3$ and selected 10,000 tokens based on TF-IDF.

\subsection{Online Catalog and Source Code}

\begin{figure}
\centering
\includegraphics[width=\columnwidth]{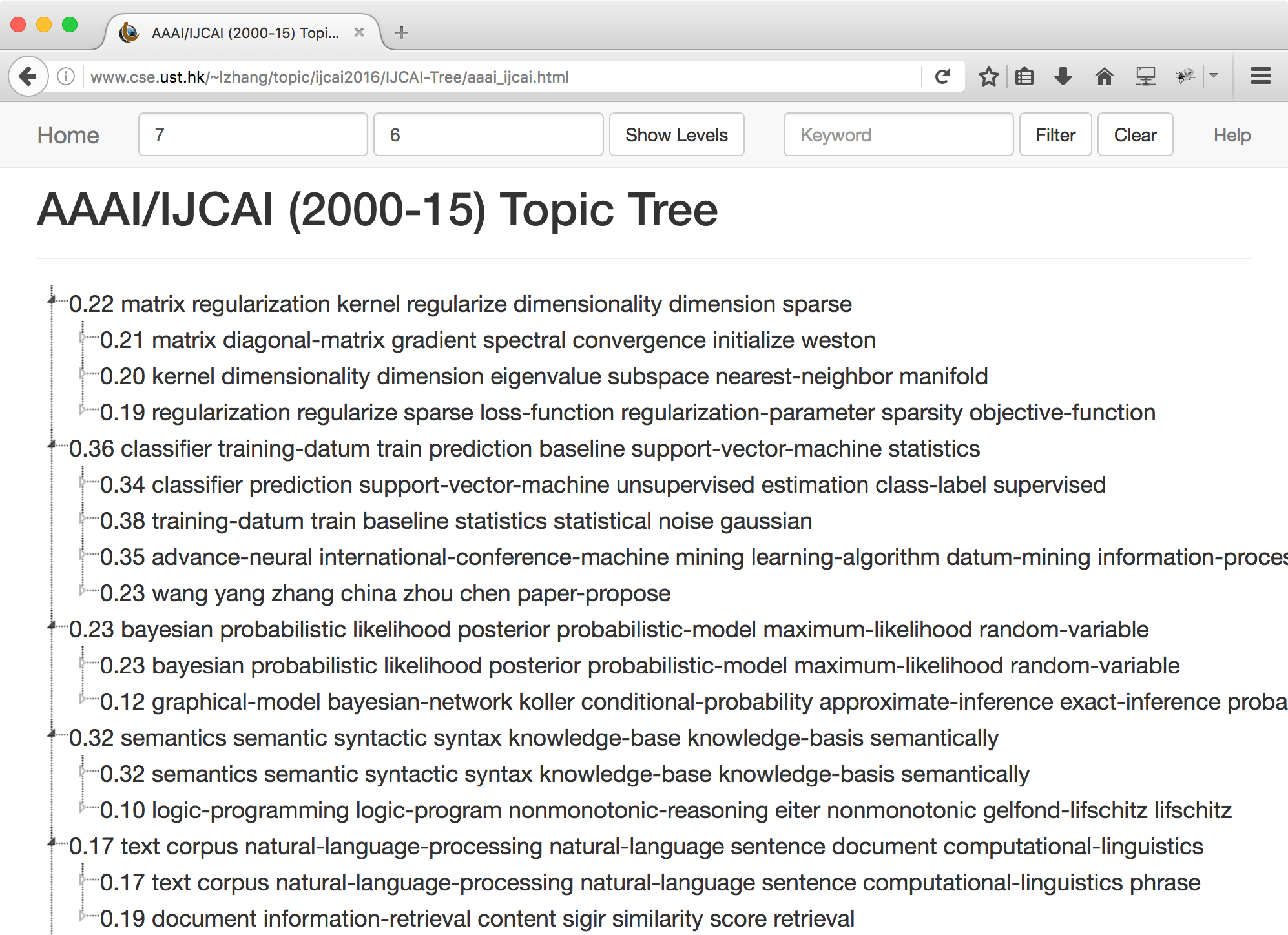}
\caption{Screenshot of online catalog.}
\label{fig:online-catalog}
\end{figure}

The online catalog can be accessed from the URL http://goo.gl/gtDJC8.  A screenshot is shown in Figure~\ref{fig:online-catalog}.  The program for building the online catalog was written in Scala and Java.  The source code can be obtained from https://github.com/kmpoon/hlta.

\subsection{Topic Hierarchy}

\begin{figure}
\centering
\includegraphics[width=\columnwidth]{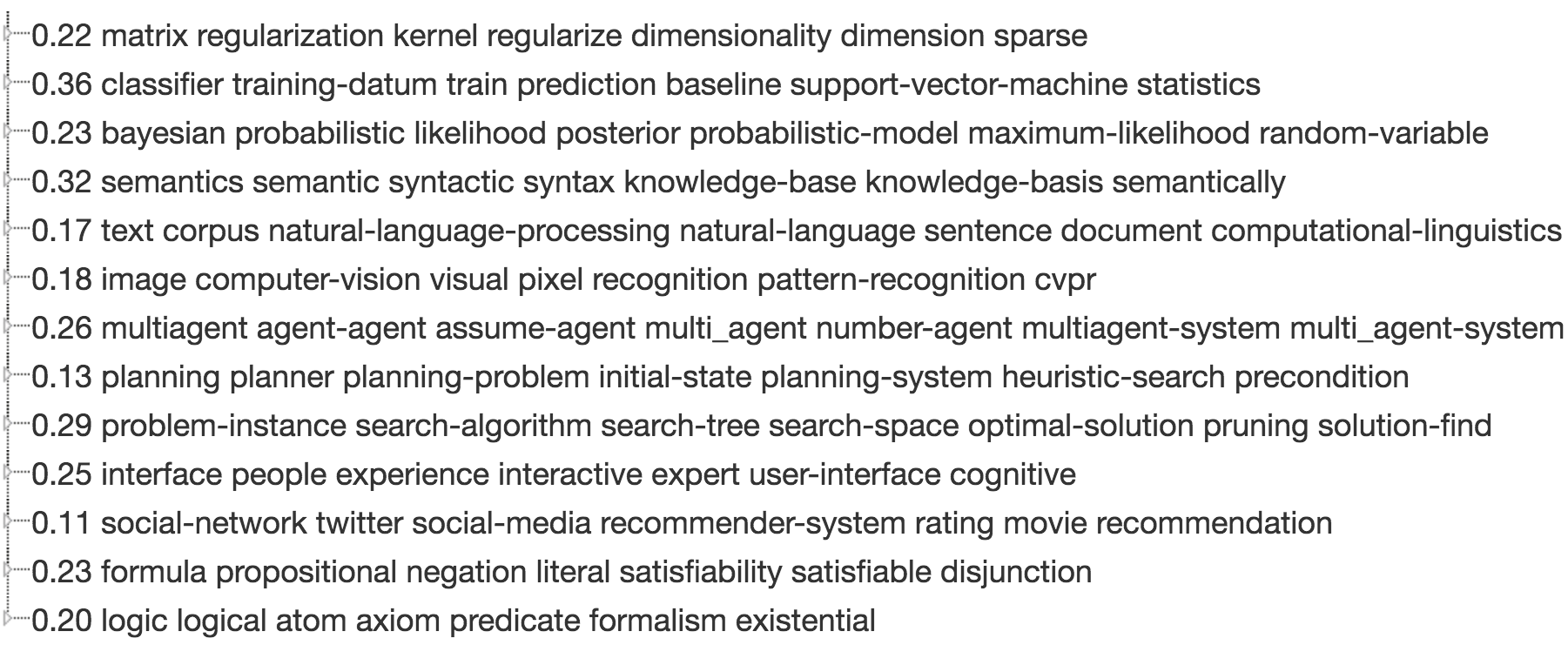}
\caption{Top level topics}
\label{fig:top-level}
\end{figure}

\begin{table}
\centering
\begin{tabular}{|c|c|} \hline
\textbf{Level} & \textbf{No. of Topics} \\ \hline
7 & 13 \\ \hline
6 & 32 \\ \hline
5 & 75 \\ \hline
4 & 179 \\ \hline
3 & 436 \\ \hline
2 & 1173 \\ \hline
1 & 3084 \\ \hline \hline
total & 4992 \\ \hline
\end{tabular}
\caption{Number of topics detected for each level.}
\label{table:levels}
\end{table}

The topic hierarchy extracted from the LTM obtained contains 7 levels of topics.  Table~\ref{table:levels} lists the number of topics for each level.  

The hierarchy contains 13 top-level topics (Figure~\ref{fig:top-level}). The 1sth topic is about kernel methods, matrix factorization, dimensionality reduction.  The 2nd one is about classifier and support vector machines.  The 3rd one is about probabilistic models and Bayesian methods.  The 4th to 7th topics are about knowledge base, natural language processing, computer vision, and agent systems, respectively.  The 8th one is about planning and heuristic search.  The 9th one is search.  The 10th one is about to people and user interface.  The 11th one is about social networks and recommender systems.  The last two are about satisfiability and logic respectively.

\subsection{Features}

The online catalog provides some useful features for topic browsing.

\paragraph{Expanding and Collapsing Topics.}  The topic hierarchy webpage allows expanding a topic node to see its child topics or collapsing a topic node to hide its child topics.  This allows users to navigate among more general topics and more specific topics.  For example, if we expand the upper 5 topics in Figure~\ref{fig:top-level}, we see some more specific topics as in Figure~\ref{fig:online-catalog}.  The bottom part of Figure~\ref{fig:online-catalog} shows that the top-level topic about natural language processing can be divided two more specific topics, one about natural language processing and one about information retrieval.

\paragraph{Levels of Topics.}  Users can specify a range of levels of topics to show at the top of the webpage (Figure~\ref{fig:online-catalog}).  By default the top two levels of topics are shown.

\begin{figure}
\centering
\includegraphics[width=\columnwidth]{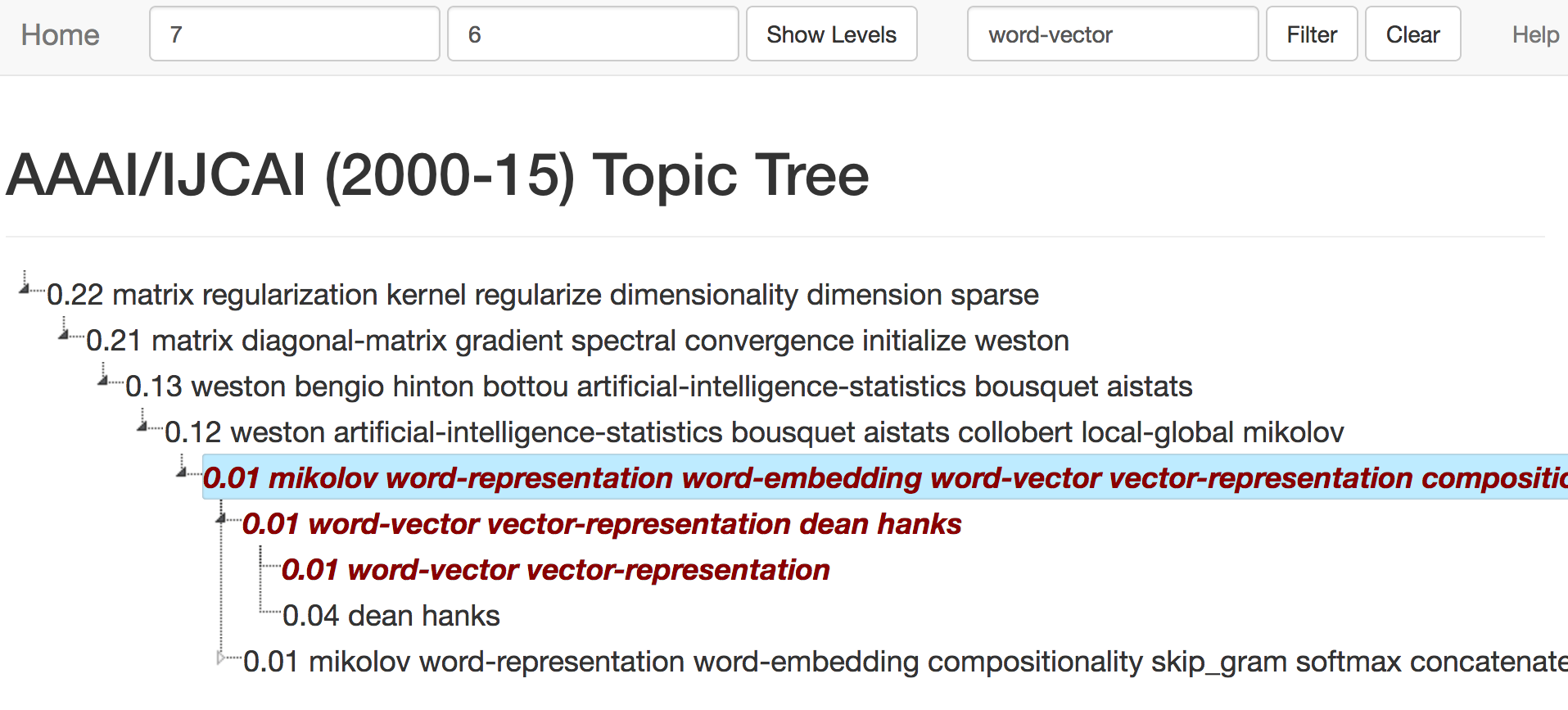}
\caption{Searching topics with keyword \word{word-vector}.}
\label{fig:search}
\end{figure}

\paragraph{Keyword Search.}  Users can enter a keyword at the top of the webpage to search for topics containing that keyword. Topics with the keyword are highlighted (Figure~\ref{fig:search}).

\begin{figure}
\centering
\includegraphics[width=\columnwidth]{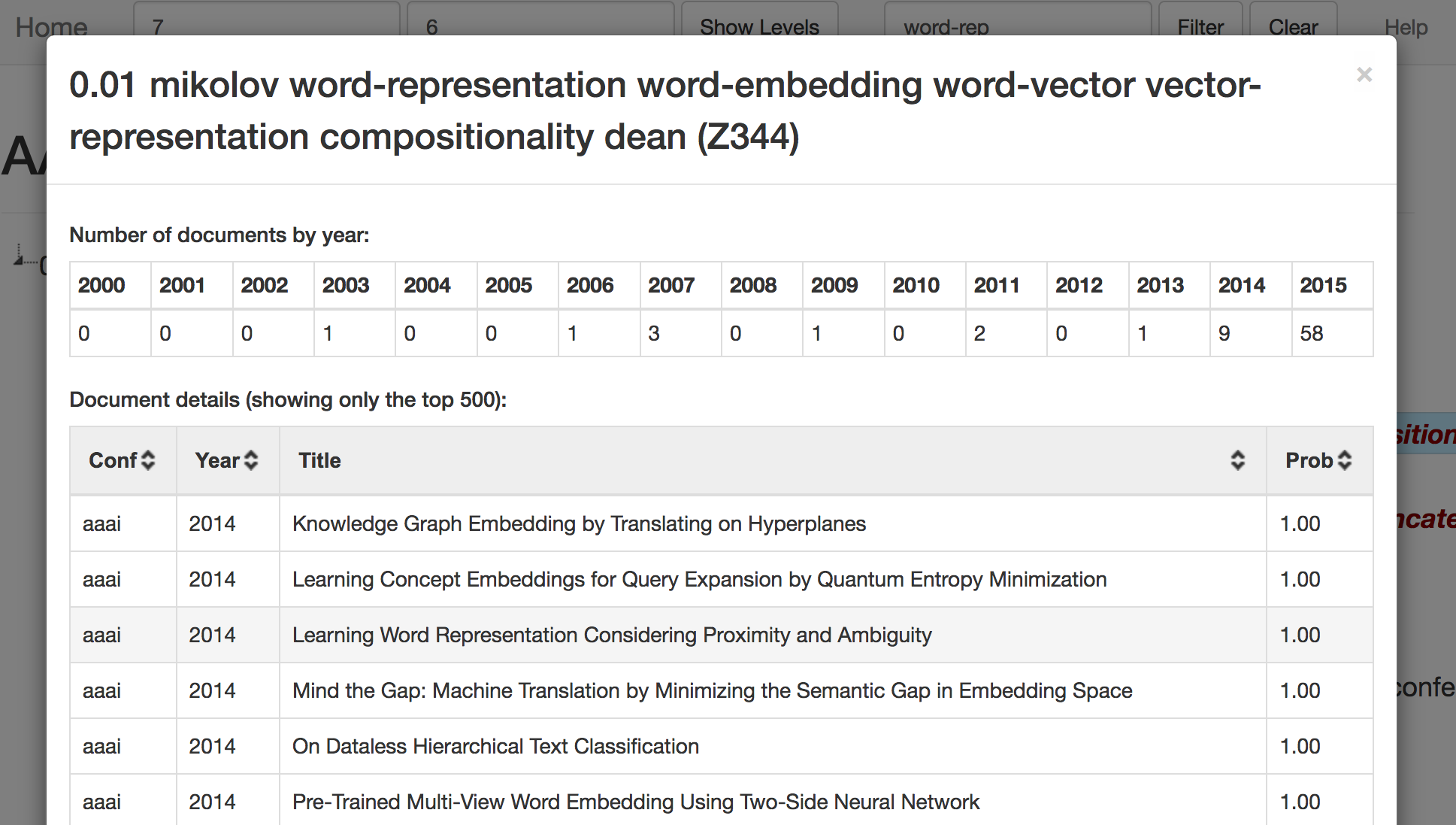}
\caption{List of papers belonging to topic $Z344$.}
\label{fig:z344-documents}
\end{figure}

\paragraph{Document List.}  A list of document belonging to a topic can be shown when a topic node is clicked (Figure~\ref{fig:z344-documents}).  The documents are sorted in descending order of membership as indicated by $P(Z|d)$.  On this page, the numbers of documents belonging to the topic for each year are also shown in a table.  We see that this topic about \word{word-representation} is emerging recently.  It has only 9 documents from 2000 to 2013 but has 9 documents in 2014 and 58 documents in 2015.

\begin{table*}
\centering \scriptsize
\begin{tabular}{cl} \hline
$Z370$ & \word{hash-function} \word{hash} \word{indyk} \word{hashing} \word{hash-method} \word{binary-code} \word{hash-code} \\ \hline
$Z344$ & \word{mikolov} \word{word-representation} \word{word-embedding} \word{word-vector} \word{vector-representation} \word{compositionality} \word{dean}\\ \hline 
$Z3207$ & \word{marketing} \word{spread} \word{viral} \word{diffusion} \word{kleinberg-tardos} \word{kempe} \word{influence-maximization} \\ \hline
\end{tabular}
\caption{Top three level-3 topics with an upward trend.}
\label{table:upward-trend}
\end{table*}

\begin{figure}
\centering
\begin{subfigure}[b]{0.32\columnwidth}
\includegraphics[width=\textwidth]{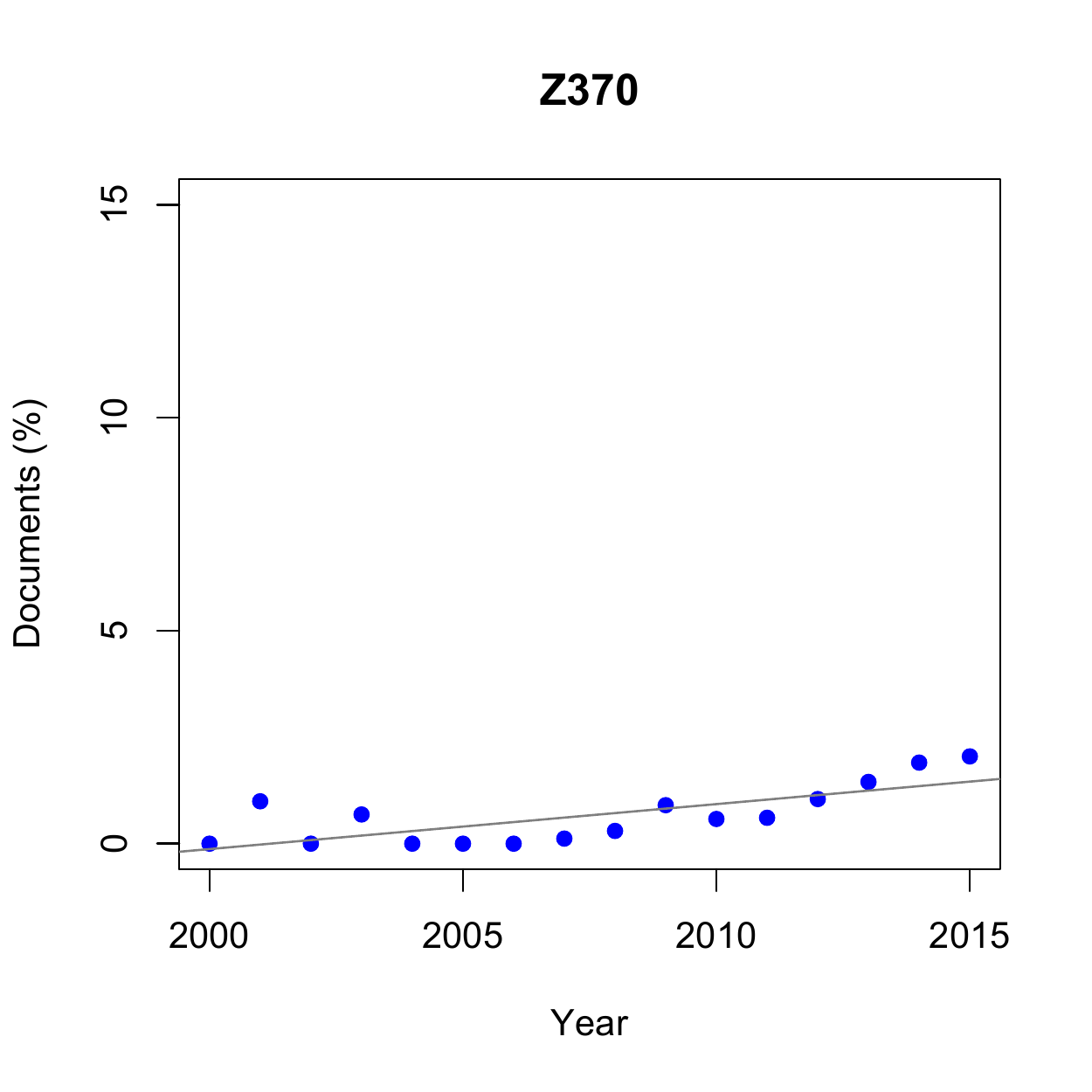}
\caption{$Z370$}
\label{fig:z370-trend}
\end{subfigure}
\begin{subfigure}[b]{0.32\columnwidth}
\includegraphics[width=\textwidth]{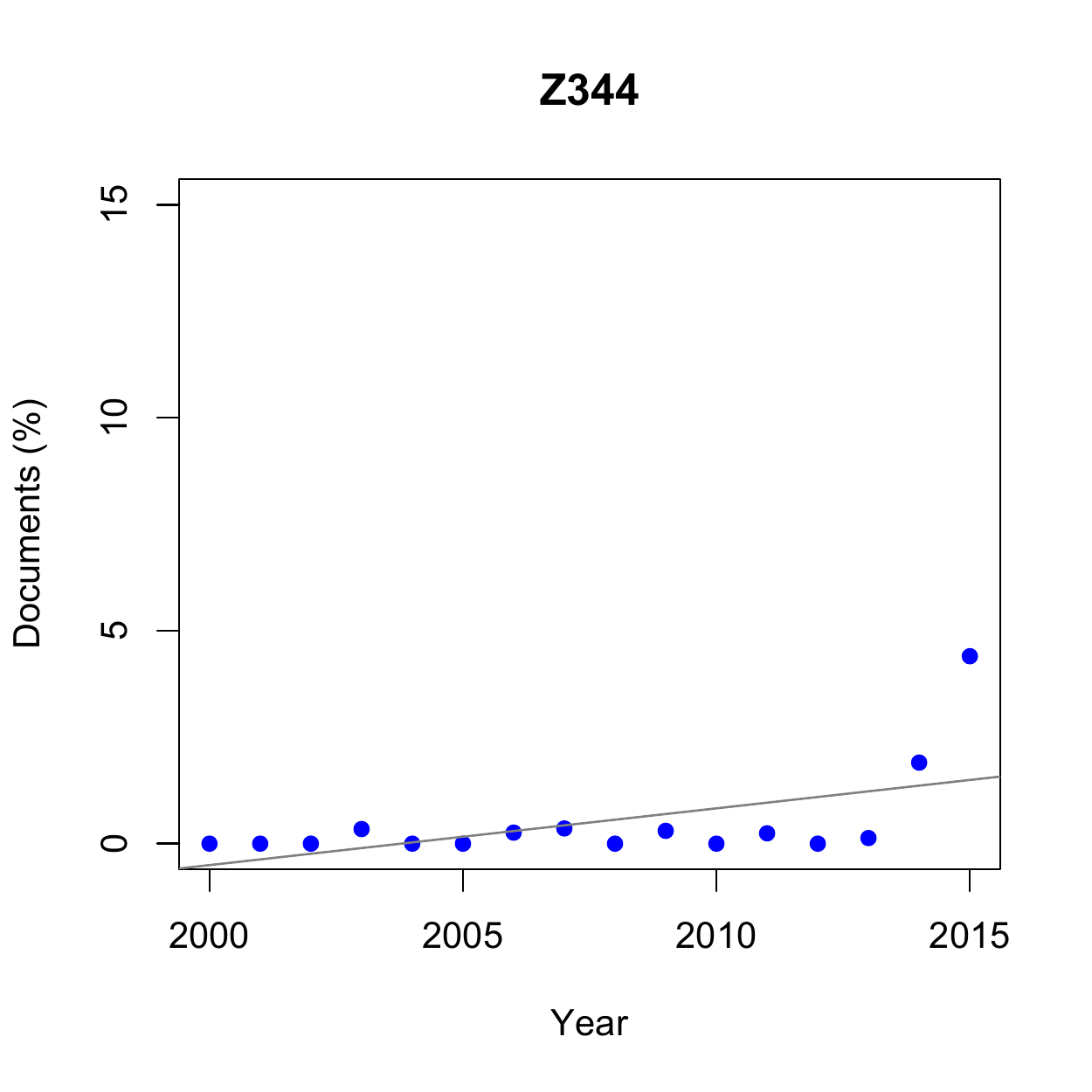}
\caption{$Z344$}
\label{fig:z344-trend}
\end{subfigure}
\begin{subfigure}[b]{0.32\columnwidth}
\includegraphics[width=\textwidth]{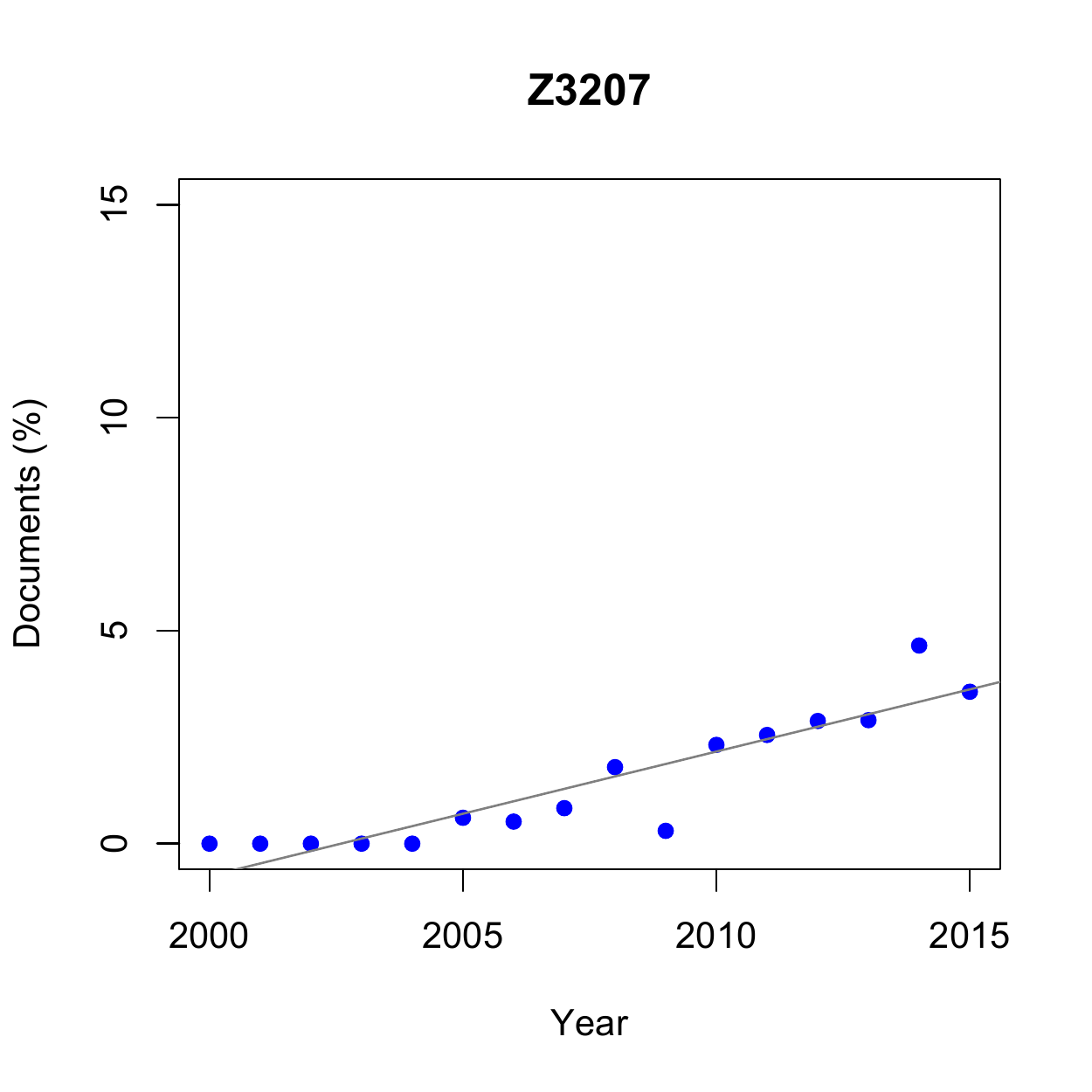}
\caption{$Z3207$}
\label{fig:z32407-trend}
\end{subfigure}
\caption{Proportion of documents belonging to a topic for each year.}
\label{fig:trend}
\end{figure}

\subsection{Trends}  
For each document $d$, we can find the year of $d$ and compute the topic indicator based on $P(Z|d)$  Hence, we can build a linear regression model using topic indicator as a predictor variable the year variable as response variable.  We can then use the regression coefficient to estimate the trend of each topic.  The trend allows researcher new to a field to consider whether a topic is worth for new work.

Table~\ref{table:upward-trend} shows the top three level-3 topics with an upward trend.  The trends are supported by the increasing proportion of documents of the topics as shown in Figure~\ref{fig:trend}.

\section{Related Works}

Topic browsing tools have been built based on topic models.  Many tools use LDA for topic modeling~\cite{gardner2010topic,chaney12visualizing,snyder13topic,sievert14ldavis}.  They do not show any hierarchy of topics.  \citet{smith14hierarchie} attempts to build a tool with a topic hierarchy by recursively splitting and remodeling a corpus based on LDA.  Unlike HLTA, the topic model does not have a strong statistical basis.

\section{Conclusions}

We present an online tool that allows users to browse topics from a hierarchy.  The topic hierarchy is built using the recently proposed \sysname{PEM-HLTA}.  The tool allows users to show documents related to each topic.  The tool provides several features for easy browsing.

In the future, we will consider using more sophisticated way to detect n-grams.  We may also include the hyperlinks to the papers in the document list for each topic.  The online catalog currently takes some time to load in a web browser.  We will consider storing the data in a database so that the loading time can be substantially reduced.

\section{Acknowledgment}

We thank Peixian Chen for the implementation of \sysname{PEM-HLTA} and Zichao Li for downloading the PDF files of the research papers.  The work in this paper was supported by the Education University of Hong Kong under project RG90/2014-2015R and Hong Kong Research Grants Council under grant 16202515.

\bibliography{us,others}
\bibliographystyle{aaai}

\end{document}